\documentclass[pmlr,onecolumn,10pt]{jmlr} 

\usepackage{booktabs}
\usepackage{siunitx}

\usepackage[switch]{lineno}

\theorembodyfont{\upshape}
\theoremheaderfont{\scshape}
\theorempostheader{:}
\theoremsep{\newline}

\jmlrvolume{LEAVE UNSET}
\jmlryear{2025}
\jmlrsubmitted{LEAVE UNSET}
\jmlrpublished{LEAVE UNSET}
\jmlrworkshop{Conference on Health, Inference, and Learning (CHIL) 2025} 

\title[Research Roundtables at CHIL 2025]{Reflections from Research Roundtables at the Conference on Health, Inference, and Learning (CHIL) 2025\\[1em]}

\author{%
\Name{Emily Alsentzer\textsuperscript{1}} \Email{ealsentzer@stanford.edu}
\AND
\Name{Marie-Laure Charpignon\textsuperscript{1}} \Email{mariecharpignon@berkeley.edu}
\AND
\Name{Bill Chen\textsuperscript{2}} \Email{bill.chen@duke.edu}
\AND
\Name{Niharika D'Souza\textsuperscript{1}} \Email{Niharika.DSouza@ibm.com}
\AND
\Name{Jason Fries\textsuperscript{1}} \Email{jason-fries@stanford.edu}
\AND
\Name{Yixing Jiang\textsuperscript{2}} \Email{jyx@stanford.edu}
\AND
\Name{Aparajita Kashyap\textsuperscript{2}} \Email{ak4885@cumc.columbia.edu}
\AND
\Name{Chanwoo Kim\textsuperscript{2}} \Email{chanwkim@cs.washington.edu}
\AND
\Name{Simon Lee\textsuperscript{2}} \Email{simonlee711@g.ucla.edu}
\AND
\Name{Aishwarya Mandyam\textsuperscript{1}} \Email{am2@stanford.edu}
\AND
\Name{Ashery Mbilinyi\textsuperscript{1}} \Email{ashery@uvic.ca}
\AND
\Name{Nikita Mehandru\textsuperscript{2}} \Email{nmehandru@berkeley.edu}
\AND
\Name{Nitish Nagesh\textsuperscript{2}} \Email{nnagesh1@uci.edu}
\AND
\Name{Brighton Nuwagira\textsuperscript{2}} \Email{brighton.nuwagira@utdallas.edu}
\AND
\Name{Emma Pierson\textsuperscript{1}} \Email{emmapierson@berkeley.edu}
\AND
\Name{Arvind Pillai\textsuperscript{2}} \Email{arvind.pillai184@gmail.com}
\AND
\Name{Akane Sano\textsuperscript{1}} \Email{akane.sano@rice.edu}
\AND
\Name{Tanveer Syeda-Mahmood\textsuperscript{1}} \Email{stf@us.ibm.com}
\AND
\Name{Shashank Yadav\textsuperscript{2}} \Email{shashank@arizona.edu}
\AND
\Name{Elias Adhanom\textsuperscript{3}} 
\AND
\Name{Muhammad Umar Afza\textsuperscript{3}} 
\AND
\Name{Amelia Archer\textsuperscript{3}} 
\AND
\Name{Suhana Bedi\textsuperscript{3}}
\AND
\Name{Vasiliki Bikia\textsuperscript{3}}
\AND
\Name{Trenton Chang\textsuperscript{3}}
\AND
\Name{George H. Chen\textsuperscript{3}} 
\AND
\Name{Winston Chen\textsuperscript{3}} 
\AND
\Name{Erica Chiang\textsuperscript{3}} 
\AND
\Name{Edward Choi\textsuperscript{3}} 
\AND
\Name{Octavia Ciora\textsuperscript{3}}
\AND
\Name{Paz Dozie-Nnamah\textsuperscript{3}} 
\AND
\Name{Shaza Elsharief\textsuperscript{3}} 
\AND
\Name{Matthew Engelhard\textsuperscript{3}} 
\AND
\Name{Ali Eshragh\textsuperscript{3}} 
\AND
\Name{Jean Feng\textsuperscript{3}} 
\AND
\Name{Josh Fessel\textsuperscript{3}}
\AND
\Name{Scott Fleming\textsuperscript{3}}
\AND
\Name{Kei Sen Fong\textsuperscript{3}} 
\AND
\Name{Thomas Frost\textsuperscript{3}}
\AND
\Name{Soham Gadgil\textsuperscript{3}}
\AND
\Name{Judy Gichoya\textsuperscript{3}}
\AND
\Name{Leeor Hershkovich\textsuperscript{3}}
\AND
\Name{Sujeong Im\textsuperscript{3}} 
\AND
\Name{Bhavya Jain\textsuperscript{3}} 
\AND
\Name{Vincent Jeanselme\textsuperscript{3}} 
\AND
\Name{Furong Jia\textsuperscript{3}} 
\AND
\Name{Qixuan (Alice) Jin\textsuperscript{3}}
\AND
\Name{Yuxuan Jin\textsuperscript{3}} 
\AND
\Name{Daniel Kapash\textsuperscript{3}}
\AND
\Name{Geetika Kapoor\textsuperscript{3}}
\AND
\Name{Behdokht Kiafar\textsuperscript{3}} 
\AND
\Name{Matthias Kleiner\textsuperscript{3}} 
\AND
\Name{Stefan Kraft\textsuperscript{3}}
\AND
\Name{Annika Kumar\textsuperscript{3}} 
\AND
\Name{Daeun Kyung\textsuperscript{3}} 
\AND
\Name{Zhongyuan Liang\textsuperscript{3}} 
\AND
\Name{Joanna Lin\textsuperscript{3}} 
\AND
\Name{Qianchu (Flora) Liu\textsuperscript{3}} 
\AND
\Name{Chang Liu\textsuperscript{3}} 
\AND
\Name{Hongzhou Luan\textsuperscript{3}} 
\AND
\Name{Chris Lunt\textsuperscript{3}} 
\AND
\Name{Leopoldo Julián Lechuga López\textsuperscript{3}}
\AND
\Name{Matthew B. A. McDermott\textsuperscript{3}} 
\AND
\Name{Shahriar Noroozizadeh\textsuperscript{3}}
\AND
\Name{Connor O'Brien\textsuperscript{3}} 
\AND
\Name{YongKyung Oh\textsuperscript{3}} 
\AND
\Name{Mixail Ota\textsuperscript{3}} 
\AND
\Name{Stephen Pfohl\textsuperscript{3}} 
\AND
\Name{Meagan Pi\textsuperscript{3}} 
\AND
\Name{Tanmoy Sarkar Pias\textsuperscript{3}} 
\AND
\Name{Emma Rocheteau\textsuperscript{3}} 
\AND
\Name{Avishaan Sethi\textsuperscript{3}}
\AND
\Name{Toru Shirakawa\textsuperscript{3}} 
\AND
\Name{Anita Silver\textsuperscript{3}} 
\AND
\Name{Neha Simha\textsuperscript{3}} 
\AND
\Name{Kamile Stankeviciute\textsuperscript{3}}
\AND
\Name{Max Sunog\textsuperscript{3}}
\AND
\Name{Peter Szolovits\textsuperscript{3}}
\AND
\Name{Shengpu Tang\textsuperscript{3}}
\AND
\Name{Jialu Tang\textsuperscript{3}}
\AND
\Name{Aaron Tierney\textsuperscript{3}} 
\AND
\Name{John Valdovinos\textsuperscript{3}}
\AND
\Name{Byron Wallace\textsuperscript{3}} 
\AND
\Name{Will Ke Wang\textsuperscript{3}} 
\AND
\Name{Peter Washington\textsuperscript{3}} 
\AND
\Name{Jeremy Weiss\textsuperscript{3}} 
\AND
\Name{Daniel Wolfe\textsuperscript{3}} 
\AND
\Name{Emily Wong\textsuperscript{3}}
\AND
\Name{Hye Sun Yun\textsuperscript{3}} 
\AND
\Name{Xiaoman Zhang\textsuperscript{3}} 
\AND
\Name{Xiao Yu Cindy Zhang\textsuperscript{3}}
\AND
\Name{Hayoung Jeong\textsuperscript{4}} \Email{hayoung.jeong@duke.edu}
\AND
\Name{Kaveri A. Thakoor\textsuperscript{4}} \Email{kat2193@columbia.edu}
}

\begin{document}

\maketitle
\footnotetext[1]{Research Roundtables Senior Chairs (authors listed alphabetically, combined with 2)}
\footnotetext[2]{Research Roundtables Junior Chairs (authors listed alphabetically, combined with 1)}
\footnotetext[3]{Research Roundtable Participants (authors listed alphabetically). Participants were offered the opportunity to elect whether or not to be listed as authors on this document.}
\footnotetext[4]{Organizing Committee for CHIL 2025 Research Roundtables (authors listed alphabetically)}

\section{Introduction}
\label{sec:intro}
The 6th annual Conference on Health, Inference, and Learning (CHIL 2025), hosted by the Association for Health Learning and Inference (AHLI), was held in person on June 25-27, 2025, at the University of California, Berkeley, in Berkeley, California, USA. As part of this year’s program, we hosted Research Roundtables to catalyze collaborative, small-group dialogue around critical, timely topics at the intersection of machine learning and healthcare.

Each roundtable was moderated by a team of senior and junior chairs who fostered open exchange, intellectual curiosity, and inclusive engagement. The sessions emphasized rigorous discussion of key challenges, creative exploration of emerging opportunities, and collective ideation toward actionable directions in the field. Overall, the Research Roundtables brought together a diverse mix of participants, including academic researchers, clinicians, industry professionals, and policy experts. 

In total, eight roundtables were held across two 30-minute sessions, with a brief transition break to allow participants to join multiple discussions. Of note, while an initial plan included a separate roundtable on \textit{Explainability/Interpretability in Sequence Models}, this was ultimately merged into a broader session on \textit{Explainability, Interpretability, and Transparency} to enable richer, more integrative dialogue.

The summaries that follow synthesize the key themes, insights, and open questions that emerged from each roundtable. Collectively, they offer a snapshot of the field’s current frontiers and a compass pointing toward its most pressing and promising future directions. 

\section{Organization Process}
The planning for the CHIL 2025 Research Roundtables began with identifying discussion topics through both retrospective review and community input. We drew from themes discussed in prior CHIL and Machine Learning for Health (ML4H) Research Roundtables, as well as trends in recent machine learning for health literature. From the collected ideas, we curated a set of four overarching themes and sub-themes:
\begin{itemize}
    \item \textbf{Explainability, Interpretability, and Transparency:} What level of explainability do clinicians and patients need to trust AI systems; Designing explainability and explanation interfaces tailored to clinicians vs. patients; Case studies of explainability techniques; Explainability/interpretability in sequence models (e.g., transformers, LLMs, etc.)
    \item \textbf{Deployment:} Ethics and Safety; Privacy; Uncertainty, bias, and fairness; Causality;  Human-computer interaction (HCI)
    \item \textbf{Scalable AI Solutions: } Synthetic Data; Learning from small medical data; Domain adaptation; Foundation models; Multimodal methods 
    \item \textbf{Scalable and Translational Health Solutions: } Integration of AI models into clinical workflows; Academic-industry partnerships; Academic-medical partnerships
\end{itemize} 

To understand the community’s interests, we launched a public call for input via a Google Form, promoted through CHIL’s website and social media platforms including Bluesky, LinkedIn, and X. In this survey, community members were asked to indicate topics they were “interested,” “less interested,” or “not interested” in, helping us prioritize the most compelling areas of discussion. We received 18 responses in total between mid-December 2024 and late February 2025. To balance breadth and depth, we ensured that at least one subtopic from each overarching theme was represented among the final roundtable sessions. The final nine topics included: \textit{``Explainability, Interpretability, and Transparency''}, \textit{``Explainability/Interpretability in Sequence Models}, \textit{``Uncertainty, Bias, and Fairness''}, \textit{``Causality''}, \textit{``Scalable AI Solutions: Domain Adaptation''}, \textit{``Scalable AI Solutions: Foundation Models''}, \textit{``Scalable AI Solutions: Learning from Small Medical Data''}, \textit{``Multimodal Methods''}, \textit{``Scalable and Translational Healthcare Solutions''}. 

For each selected topic, we invited one senior chair and at least one junior chair with relevant domain expertise to moderate the discussion. We prioritized individuals who had already expressed interest in participating either by completing the volunteer form circulated alongside the topic interest survey or by submitting a full paper or doctoral symposium contribution relevant to the roundtable themes. In addition, members of the program committee were encouraged to invite local researchers and practitioners to serve as chairs.

In preparation for the roundtables, each chairing team collaboratively developed a set of 3–5 guiding prompts (e.g., discussion questions to hypothetical case studies) which were shared publicly on the CHIL 2025 conference website two weeks prior to the event. Following the conference, each team synthesized their discussion into a written summary, capturing key insights, open questions, and reflections from their table. The summaries presented here reflect a partial synthesis of the conversations held during each roundtable. They are not intended as endorsements or official positions of any individual participant, and may include ideas or claims with which some attendees disagree.

\section{Explainability, Interpretability, and Transparency}
\label{sec:table1&2}
\paragraph{Subtopic:} What types of explanations do different healthcare stakeholders—clinicians, patients, and developers—actually need, and how can we design user-centered explainability systems to meet those needs? What are the practical limitations of current explanation methods when deployed in real clinical environments, and how can we overcome them? How can explainability techniques be adapted to temporal and longitudinal healthcare data, where traditional attribution methods often fall short?

\paragraph{Chairs:} Ashery Mbilinyi, Tanveer Syeda-Mahmood, Chanwoo Kim, Shashank Yadav, and Nikita Mehandru

\paragraph{Participants:} Matthew Engelhard, Kei Sen Fong, Soham Gadgil, Stefan Kraft, Meagan Pi, Jialu Tang, Kaveri Thakoor, and Byron Wallace

\subsection{Background} 
The demand for explainable AI in healthcare has increased as machine learning models have become increasingly complex and prevalent in medicine. The emergence of foundation models and large language models in healthcare has highlighted new risks, particularly around hallucinations and prompt sensitivity. 

Numerous explainable AI methods have been proposed to address these challenges. These methods aim to make the predictions of complex machine learning models more transparent, and understandable to end users. In the case of black-box models, such as deep neural networks, explanations are typically generated post-hoc using techniques like SHAP \citep{lundberg2017} or LIME \citep{ribeiro2016}, which attribute model outputs to input features. The overarching goal of explainability is to provide meaningful rationales behind model decisions to enhance trust, support verification, and enable safe and effective use in high-stakes settings like healthcare.

Despite this methodological progress, many explainability techniques fail to gain traction in clinical practice—not due to algorithmic limitations, but because of challenges related to workflow disruption, integration into clinical systems, and misalignment with clinicians’ mental models. This has led to growing recognition that explainability is not solely a technical problem, but a human-centered design challenge that demands a deep understanding of stakeholder needs and real-world constraints.

\subsection{Discussion} Over ten attendees participated in the CHIL roundtable on \textit{Explainability, Interpretability, and Transparency}. Participants included academic researchers developing explainability methods as well as industry professionals working on real-world deployment. Key themes from the discussion are summarized below.

\subsubsection{User-centered Explainability in Practice}
A central and recurring theme present in our discussion was the critical need for explainability methods to be tailored to the user's background and goals. Clinicians, for instance, demand explanations that are concise, actionable, and intuitively aligned with their preexisting mental models and medical training. Clinicians typically reason through high-level clinical abstractions such as specific symptoms, diagnoses, pathophysiological processes, and disease stages. Tools resembling familiar formats, such as clinical scoring tables or ``MD Calc''-style outputs, often prove more effective than highly technical analysis, because they integrate seamlessly into existing clinical workflows. For patients, it’s crucial to translate complex health information into accessible language, especially where low levels of medical literacy exist, so that they’re empowered to more actively engage in their own care (e.g., using continuous glucose monitor devices). A promising design paradigm involves offering a simple, high-level summary with the option for users to ``drill down'' into more detailed or technical explanations on demand, effectively managing cognitive load and catering to varied transparency needs. The conversation highlighted that successful adoption of AI depends more on trust and usability than technical performance. 

\subsubsection{Practical Limitations of Existing Explanation Methods}
The practical deployment of explainable AI in healthcare environments faces several hurdles. First, post-hoc explanation methods such as SHAP and LIME are computationally intensive and slow, limiting their use in real-time settings. As tools such as \href{https://www.openevidence.com/}{OpenEvidence}, which provides LLM-powered research summaries for clinicians, are rapidly deployed, concerns around reliability in medical settings have also taken center stage. These include extreme prompt sensitivity, where subtle changes in question phrasing can elicit entirely contradictory medical advice, and hallucinations. The discussion underscored the urgent need for robust validation, oversight mechanisms, and improved transparency.

\subsubsection{Implementation Challenges}
The successful adoption of explainable AI tools in healthcare also hinges on how seamlessly these tools integrate into clinical workflows. Participants emphasized that many AI systems fail because they introduce friction—requiring clinicians to interact with new dashboards, deal with pop-ups, or switch contexts—thereby increasing cognitive load and disrupting routine care. Embedding AI explanations directly into existing electronic medical records (EMRs) was highlighted as a more effective strategy. However, deep integration with platforms like Epic remains a significant barrier. These systems are often closed and require months of backend development, data warehousing, and regulatory approvals. Even promising tools can stall due to these hurdles, delaying their business value and clinical impact. These infrastructural and organizational hurdles were seen as significant bottlenecks to translation and impact.

\subsubsection{Education}
There was broad agreement that widespread adoption of explainable AI requires not just technical innovation but cultural transformation. Historical precedents, such as IBM’s radiology AI demonstrations at RSNA 2016 \citep{ibmresearch2017}, show that clinician involvement and targeted education can foster trust and engagement. Participants stressed that trust must be earned through iterative design, transparency, and sustained collaboration. 

\subsubsection{Explainability in the Time-Series Data}
Explainability in time-series and longitudinal clinical data presents distinct and complex challenges due to the temporal dependencies. In settings such as ICU-monitoring or chronic disease management, model decisions must be contextualized within a patient’s evolving history, including interventions and symptom progression. Traditional attribution methods like gradients, SHAP, or LIME often fall short, as they struggle to maintain temporal causality or capture the sequential relationships between events. Participants emphasized the need for approaches that align better with clinical reasoning, such as episodic memory models and event-based segmentation, which partition continuous data into clinically meaningful episodes. These techniques aim to provide explanations that are temporally coherent, clinically valid, and intuitive to users. By incorporating the structure of real medical timelines, time-aware models can improve trust and utility, making them more suitable for high-stakes applications in healthcare. The discussion highlighted that clinicians require explanations that explicitly reflect how patient trajectories evolve over time. Explanations should clearly indicate why a patient’s risk status changed at specific moments and should be directly linked to physiological changes, treatments administered or disease progression.

\section{Uncertainty, bias, and fairness}
\label{sec:table3}
\paragraph{Subtopic:} When incorporating machine learning into health contexts, it is imperative to understand how our choices impact individuals and populations, particularly those who experience sociopolitical barriers to receiving high-quality healthcare. What recommendations can we make for unbiased and safe use of AI in healthcare? How can we appropriately communicate bias, fairness, and uncertainty considerations with different healthcare stakeholders?

\paragraph{Chairs:} Emma Pierson and Aparajita Kashyap 

\paragraph{Participants:} Trenton Chang, Erica Chiang, Paz Dozie-Nnamah, Jean Feng, Vincent Jeanselme, Qixuan (Alice) Jin, Matthias Kleiner, Leopoldo Julián Lechuga López, Stephen Pfohl, Tanmoy Sarkar Pias, Max Sunog, Daniel Wolfe, Emily Wong, and Hye Sun Yun

\subsection{Background} In the AI/ML literature, there is a large body of research devoted to assessing and improving bias, uncertainty, and fairness \citep{lopez2025,mehrabi2021}. Exploration of these topics is critical for improving the robustness, reliability, and generalizability of AI/ML models across all settings. During this roundtable discussion, we aimed to discuss new questions of uncertainty, bias, and fairness that have risen in AI/ML (particularly around LLMs, data generation, and the machine learning development pipeline) and examine them within the specific context of health and healthcare. 

\subsection{Discussion} During group discussions, three recurring recommendations arose: first, consider fairness within the context of the specific task and setting; second, incorporate fairness assessments and interventions at all points in the AI/ML pipeline; and third, collaborate more with patients and clinicians. The group discussed several use cases related to these recommendations, but all three ideas were directly relevant when examining the design, evaluation, and deployment of LLMs. 

\subsubsection{Context and task-specific considerations of fairness}
During model development and initial evaluation, there is a range of metrics that one can use to assess model fairness. It is often impossible to simultaneously equalize multiple metrics across groups \citep{kleinberg2016, chouldechova2016}, and optimizing for inappropriate metrics in the name of fairness can cause harm \citep{pfohl2021}, making it imperative to choose metrics appropriate to the task at hand. Fairness in the context of AI-generated patient messages should be measured differently from fairness in the context of a disease risk prediction task, which should in turn be measured differently from fairness of a resource allocation task.

Decision theoretic approaches may be useful for implementing fair models, as they more explicitly consider the impacts of the model post-deployment \citep{scantamburlo2025}. Model calibration is also important for ensuring clinical translation and utility \citep{walsh2017}. When measuring model fairness, common practice involves measuring subpopulation performance on disaggregated groups. However, this may prove unreliable given certain data-generating processes, and should be augmented with other approaches where appropriate \citep{pfohl2025}.

\subsubsection{Addressing fairness throughout the AI/ML pipeline}
In line with \citet{chen2021}, the group discussed the importance of making choices to promote fairness and equity throughout the research process. 

One critical choice, which is often under-considered by AI researchers, is what problem to tackle in the first place. In choosing a problem, it is important to acknowledge the myriad of healthcare problems that require increased social and structural support (e.g. moving to housing in a less polluted area or addressing air pollution is likely to be a more impactful intervention for asthma than any machine learning model \citep{grant2022} and focus on use cases where an algorithmic intervention could improve health disparities or increase access to care is likely to be useful. In addition to the problem itself, the problem setting is an important consideration: most AI interventions are developed at large urban academic health centers (and in fact, the majority of this roundtable group belong to such institutions), and development and validation of AI tools in rural and under-resourced health settings remains an open question. 

It is well-known that healthcare data generally reflects the biases and disparities of the system in which it was recorded. Publicly available healthcare datasets have traditionally lacked diversity \citep{arora2023}, but newer datasets (e.g. \href{https://docs.ngsci.org/}{Nightingale}, \href{https://allofus.nih.gov/}{\textit{All of Us}}) are prioritizing diversity and representation. Additionally, including or excluding certain medical codes during phenotyping may drastically change the demographic makeup of a cohort \citep{sun2023}, and some individuals may not be able to receive formal diagnoses at all. A patient’s outcomes are also reflective of potentially biased human decisions. While there are some ways to address biases in the data collection practices (e.g. screening everyone rather than only “at-risk” individuals), this is not feasible (and may not be appropriate) across all conditions or all clinical settings. The field thus requires more methodological research around characterizing and accounting for bias in the data-generating process specifically \citep{chiang2025, shanmugam2023}.

\subsubsection{Inclusion of patient and clinician stakeholders}
Despite the large volume of AI interventions that seek to improve patient care and outcomes, relatively few of them are developed in collaboration with clinicians, and still fewer are developed in collaboration with patients \citep{tulkjesso2022, moy2024}. The group suggested that patients (and providers) should be invited into conversations as early in the development process as possible, including during the problem selection and research question refinement phase. Understanding how patients want to get involved, what kinds of data they feel comfortable with a model using, and how they conceptualize algorithmic fairness are all valuable insights that can inform problem definition, data collection, and model design.

When considering model evaluation, consulting with patients and clinicians may inform appropriate choices of metrics, particularly in cases where there is no single “best metric”. For example, in the case of classification, deciding on an appropriate fairness metric may require weighting the impact of a false positive versus a false negative prediction, and contextualizing these incorrect predictions within any known health disparities surrounding the use case. Consulting a diverse set of stakeholders can help developers identify and prioritize such considerations.

Successful models of patient collaboration include the Patient-Centered Outcomes Research Institute (\href{https://www.pcori.org/}{PCORI}), which prioritizes stakeholder engagement in research and the Treatment Action Group (\href{https://www.treatmentactiongroup.org/about-us/history/}{TAG}), which advocated for fast, high-quality clinical trials for HIV/AIDS medication. Given that people’s openness to AI is in part dependent on their AI literacy \citep{bewersdorff2025}, increasing patient literacy around AI will be an important step in starting productive conversations about AI in healthcare. 

\subsubsection{Safety and fairness of LLMs}
All three of the previously discussed principles proved highly relevant to LLMs. During the group discussion, LLMs were broadly noted as having potential to improve access to healthcare \citep{pierson2025}: for example, they can be used to translate jargon-filled consent forms, or can be used by individuals who are unable to access or afford in-person care. For rare diseases in particular, LLMs can help patients make sense of their complex symptoms \citep{ao2025}. However, there are concerns about the extent to which reliance on LLMs to fill in the well-documented gaps of healthcare access may heighten disparities in access to high-quality healthcare. It is important to ensure that expanding use of LLMs in clinical settings does not come at the expense of expanding face-to-face access with clinicians. Additionally, if people are using LLMs because they don’t have access to a clinician, they likely have no way to verify its output (which most LLMs recommend when dispensing medical information). Improving AI and technical literacy to make LLMs more accessible to the general public will also be crucial to meaningfully improving healthcare access using LLMs. Additionally, there are open questions around the relationship between trust in LLMs and trust in healthcare providers: to what extent are these quantities correlated? Can LLMs help people escalate care when appropriate?

One key barrier to safe use of LLMs is our inability to evaluate unstructured LLM output at scale, particularly when evaluating for bias and fairness, although valuable new resources are being developed in this regard \citep{pfohl2024}. Quality is a multidimensional, complex, and highly context-dependent construct; however, most evaluations of LLMs in healthcare settings are restricted to a narrow band of tasks and medical specialties \citep{bedi2025}. Further work is needed to develop frameworks for assessing LLM output quality across a wider set of use cases, particularly those that reflect how patients (rather than providers) leverage LLMs. Using an LLM as a judge can help evaluate large volumes of LLM output \citep{arora2023}, and may help under-resourced hospital systems that lack resources for human annotation. However, this approach may miss problems with bias and unfairness \citep{ye2025}. Collaboration with clinical and non-clinical professionals (e.g. ethicists) was named as a valuable resource for LLM evaluation, as they could conduct more rigorous and nuanced content analyses of LLM output. When working with expert collaborators, however, it is important to not rely solely on interrater reliability as a measure of evaluation quality, as people may capture different nuances informed by their different specialties or approaches to science.

\section{Causality}
\label{sec:table4}

\paragraph{Chairs:} Marie-Laure Charpignon and Nitish Nagesh 

\paragraph{Participants:} Winston Chen, Octavia Ciora, Thomas Frost, Hayoung Jeong, Shahriar Noroozizadeh, Emma Rocheteau, Toru Shirakawa, Peter Szolovits, and Will Ke Wang

\subsection{Background} Our causality roundtable gathered both junior and senior researchers, spanning multiple domains—from computer science and informatics, to statistics and epidemiology, to medicine, and to public policy. The objective was to learn from each other’s expertise, share about our ongoing projects, and make meaningful connections that will last for years.

\subsection{Discussion} Attendees of the causality roundtable included doctoral students and postdoctoral fellows, physician-scientists, and professors.We focused the first part of the roundtable discussion on causal discovery in high-dimensional health datasets (e.g., electronic health records, physiological signals and logs from wearables) and kicked off the conversation with the following questions: 
\begin{enumerate}
    \item How can we sample large-scale datasets to learn causal graphs/diagrams?
    \item How can we evaluate learned causal graphs/diagrams in the absence of their ground-truth version?
    \item How can we run rigorous validation studies with domain experts (e.g., clinicians, epidemiologists)? What are good ways to facilitate their interactions with potentially large causal graphs/diagrams learned automatically from data?

\end{enumerate}
The beginning of the conversation centered around the use of large language models (LLM) in causal inference. For example, since most applied causal inference projects start with drawing a causal graph/diagram or DAG, the use of LLMs could either replace, supplement, or complement discussions with domain experts to ultimately enhance both the completeness and robustness of the graph/diagram/DAG. Indeed, human experts who specialize in a given public health and medicine domain (e.g., sepsis, dementia) constitute a scarce resource. In this context, one could thus envision adopting a Bayesian approach, i.e., first collecting a few causal graphs/diagrams drawn by/from domain experts, using them as seeds or priors, and subsequently learning a more comprehensive DAG from large-scale data such as electronic health records, claims, or logs derived from patient activities or even multiple eligible DAGs/or even a class of eligible DAGs.

As a group, we discussed the possibility of relying on LLMs to enrich the set of features to be considered for inclusion in a propensity score model for treatment assignment, as screening the literature at scale may facilitate the elicitation of weak confounders that a human expert might have neglected at first. This candidate set of confounders would then be validated with one or several human experts, prior to model implementation.

Relatedly, we briefly discussed a recent paper contributed by Irene Y. Chen and members of the CHIL community on the use of LLMs to better understand the reasons underlying contraceptive switches \citep{miao2025a}.

The physician-scientists at our table reminded the group of opportunities to interrogate the safety and effectiveness of allocated treatments—even in settings where protocols exist (e.g., time to initiate mechanical ventilation for an ICU patient, type of antibiotics to administer while waiting for microbiology results). As candidly exclaimed, physicians could be thought of as  “noisy decision-makers”. Indeed, decisions can be fuzzy in clinical practice, although reference/recommended thresholds have been established (e.g., for ordering a blood transfusion or allocating an anti-arrhythmic treatment). In addition, clinical practice patterns and guidelines may vary across health systems, medical societies, and countries. For example, cardiologists and primary care physicians are generally given the discretion to prescribe either an ACE inhibitor or an angiotensin receptor blocker to a patient recently diagnosed with hypertension. Further, the European Society of Cardiology recommends providing antihypertensive treatment to patients with a systolic blood pressure greater than 140, while the American College of Cardiology/American Heart Association uses a cut-off of 130. In sum, differences in the implementation of clinical guidelines and/or delays in the medical decision-making process can induce small deviations from standardized protocols – such sources of heterogeneity are critical to identify and highly valuable to draw inferences about what works.

The group also discussed the importance of connecting the fields of causal inference and reinforcement learning for more pragmatic and trustworthy treatment effect estimation. For example, knowledge about the patient’s medication adherence over time shall/should be incorporated in survival analyses aimed at evaluating the comparative effectiveness of sustained treatment strategies for chronic disease management, especially when there is evidence that clinicians update their prescriptions based on the patient’s health awareness and compliance.

Our group also discussed the growing field of research on when and how to best combine data from randomized controlled trials (RCT) and observational studies (e.g., based on electronic health records and/or claims data). Notable reviews on that topic include \citet{dahabreh2024}, \citet{rosenman2025} and \citet{colnet2023}.

Relatedly, the group underscored the importance of assessing the generalizability and transportability of causal effect estimates, noting that some causal estimands may be easier to transport than others. Important reviews and studies covering these questions include \citet{degtiar2023}, \citet{huang2024a}, \citet{colnet2024}, \citet{manke-reimers2025}, \citet{boughdiri2025}, and \citet{even2025}. 

Finally, our group discussed the need for our community to interrogate the use of methods emanating from the economics/econometrics literature and possibly adopt them more widely, when relevant. Such methods include quasi-experimental methods (e.g., regression discontinuity designs around official medical thresholds), synthetic control approaches (e.g., to assess the effect of a change in medication availability, clinical score definition, or treatment allocation policy), as well as learning strategies based on instrumental variables (e.g., provider prescription preferences).

\section{Domain Adaptation}
\label{sec:table5}

\paragraph{Subtopic:} Domain adaptation remains a critical step in enabling the wide deployment of machine learning or artificial intelligence models. Through much of the conference, individuals raised concerns about the ability for machine learning models and LLMs to generalize to populations presumably unseen in their corresponding training datasets. The goal of this discussion was to discuss ways to better enable domain adaptation through algorithmic strategies and different ways to mitigate the effects of small datasets. 

\paragraph{Chairs:} Aishwarya Mandyam and Simon Lee
\paragraph{Participants:} Elias Adhanom, George H. Chen, Ali Eshragh, Sujeong Im, Geetika Kapoor, YongKyung Oh, and Kaveri Thakoor

\subsection{Background} 
Domain adaptation encompasses a number of problems. Within the context of machine learning for healthcare, domain adaptation mainly refers to the problem of training a machine learning model using data from one dataset/patient distribution, and then deploying it with the expectation that the test time patient distribution may be distinct. This presents several challenges, particularly in terms of bias. However, there have been many broad efforts to mitigate the risks from domain adaptation such as gathering larger datasets, as well as algorithmic advances. Recently, Large Language Models (LLMs) are starting to get deployed widely, and identifying ways to responsibly deploy models given the dynamics at play remains an open challenge.

\subsection{Discussion} The discussion at the roundtable can be broadly split into two categories: the curation of datasets, and possible algorithmic choices to enable better domain adaptation. 

\subsubsection{Dataset Curation}
To improve domain adaptation in machine learning for healthcare, significant changes to the way datasets are gathered and constructed are necessary. One key challenge is that most large-scale healthcare datasets used in training do not reflect the specific characteristics of individual hospitals or patient populations. This mismatch limits model generalizability and poses risks when deploying models in new environments. To address this, future datasets should aim to capture broader demographic, geographic, and clinical diversity. This could involve creating collaborative data-sharing frameworks across multiple hospital systems, which, although difficult due to regulatory and privacy concerns, would provide the heterogeneity needed for models to generalize more effectively. Federated learning and secure aggregation techniques could support such initiatives by enabling shared learning without centralized data pooling.

In addition to broader real-world data collection, synthetic data offers a promising avenue for augmenting datasets to support domain adaptation. Carefully designed synthetic data, particularly when it reflects the distributional characteristics of underrepresented populations or rare clinical events, can help mitigate dataset imbalances. This includes upsampling strategies and simulation-based approaches that can enrich smaller datasets such as those containing information for a specific hospital system. However, it's important to distinguish between synthetic and purely simulated data, as the former can be guided by real data distributions and used to supplement training in more targeted ways. Further, models should be built with mechanisms such as uncertainty quantification or active learning that can leverage these enriched datasets to identify when adaptation is needed and guide further data collection or fine-tuning accordingly.

\subsubsection{Algorithmic Choices}
To support domain adaptation in healthcare machine learning, algorithmic choices play a crucial role alongside dataset improvements. One promising direction is the use of active learning, where models can identify and request the most informative examples for labeling. This could be especially useful in healthcare settings with limited annotated data or when adapting to new hospital environments with distinct patient populations. By focusing labeling efforts on uncertain or representative cases, active learning can improve sample efficiency and help models adapt more quickly to distributional shifts. However, realizing the potential of active learning requires better infrastructure and workflows within hospital systems to support real-time data acquisition and model updates.
Another key algorithmic strategy is instruction tuning, which can help LLMs adapt more effectively to domain-specific tasks by guiding them with examples or task-specific instructions. Moving away from purely zero-shot paradigms and instead allowing LLMs to expand their responses and express uncertainty can also be valuable, especially in high-stakes healthcare contexts. These approaches may enable models to better recognize what they don’t know and hedge their outputs accordingly, a capability that is crucial when models are deployed in unfamiliar environments. Furthermore, rather than relying solely on general-purpose architectures like transformers, there is an opportunity to design algorithms that explicitly exploit the structure and constraints of healthcare data—such as temporal dynamics, hierarchical codes, or missingness patterns—leading to more robust domain adaptation. Ultimately, algorithmic advances that integrate learning under uncertainty, personalization, and domain-specific inductive biases will be key to safe and effective healthcare deployment.

\section{Foundation Models}
\label{sec:table6}

\paragraph{Chairs:} Akane Sano and Arvind Pillai 

\paragraph{Participants:} Edward Choi, Scott Fleming, Furong Jia, Daniel Kapash, Qianchu (Flora) Liu, Matthew B. A. McDermott, Mixail Ota, Anita Silver, John Valdovinos, Will Ke Wang, Jeremy Weiss, and Xiaoman Zhang

\subsection{Background} Foundation models, large-scale, pre-trained AI architectures capable of generalizing across tasks, have a potential to reshape biomedical research and clinical care. These models trained on diverse and large data including electronic health records, imaging, clinical notes, genomic data, and wearable data can serve as a cornerstone for supporting diagnosis, personalized treatment and uncovering insights from complex datasets. Our roundtable explored several critical questions concerning foundation models: (1) integration of foundation models into clinical workflows, (2) adaptability to distribution shifts, and (3) integration of multimodal data and collaborative framework.

\subsection{Discussion}

\subsubsection{Integrating foundation models in clinical workflows}
The initial discussion at the roundtable highlighted the challenges of integrating health foundation models into clinical workflows. Participants generally agreed that current knowledge about these models is insufficient for effective clinical integration. Specifically, topics outside of machine learning, such as ethics, economics, and policy, were recurring themes. An interesting example surfaced regarding the use of Large Language Models (LLMs) in clinical settings. While EHR system companies such as EPIC started to integrate LLMs in their systems, they have not been fully integrated in the majority of clinical flows.  The group pointed out that some doctors are, in fact, using various generative AI tools for brainstorming medical solutions.

Participants discussed the importance of human factors (e.g., low cost maintenance, trust, seamless integration and minimum interference, perceived values). One example discussed was an assistant foundation model that processes EHR context and doctor–patient conversations to generate suggestions without disrupting established clinical routines.

\subsubsection{Adapting foundation models to dynamic environments}
Participants next addressed how foundation models can remain robust amid data distribution shifts and heterogeneous populations. Traditional domain adaptation and domain generalization techniques often underperform in the clinical domain compared to other areas such as computer vision. Often fine-tuning foundation models yields better results. One important reason discussed is that real-world EHRs exhibit a significant distribution shift, not only in data formats but in how diseases are defined across institutions. As a solution, the participants emphasized regular model monitoring and proposed exploring online learning and continual learning to address data shifts. 

\subsubsection{Multimodal foundation models}
The final discussion focused on multi-modal foundation models. One promising direction is a modular, agent-based framework, where multiple specialized foundation models coordinate and perform reasoning. Such a modular framework will allow people to use only the models required for their analysis rather than one large FM. Participants predicted that model composition protocols (MCPs) and agent systems might lead research directions in the coming years.

\section{Learning from small medical data }
\label{sec:table7}

\paragraph{Chairs:} Emily Alsentzer and Brighton Nuwagira

\paragraph{Participants:} Muhammad Umar Afza, George Chen, Bhavya Jain, Yuxuan Jin, Joanna Lin, Chang Liu, Hongzhou Luan, Avishaan Sethi, Neha Simha, Xiao Yu, and Cindy Zhang

\subsection{Background} Many real-world healthcare AI applications are constrained not by model design but by access to high-quality data. The roundtable explored how to overcome this limitation through principled use of synthetic data, federated learning, domain knowledge incorporation, and evaluation strategies. The discussion spanned technical, ethical, and operational barriers to learning from small datasets and achieving generalizability without centralized data sharing.

\subsection{Discussion}
The discussion brought together diverse perspectives from researchers, clinicians, and engineers, focusing on both methodological and systemic approaches to learning from limited healthcare data. The conversation covered five central themes: the promises and challenges of synthetic data, the role of domain knowledge in enhancing model robustness, strategies for federated and collaborative learning, robust evaluation in small data settings, and institutional and policy barriers to data sharing and innovation.

\subsubsection{Synthetic Data}
Participants expressed a range of opinions about the viability of synthetic data. Structured synthetic data generation methods—such as those guided by biomedical ontologies or clinical heuristics—were viewed more favorably than unconstrained GANs, particularly for small, rare disease cohorts. Validation challenges included maintaining diversity and avoiding hidden artifacts. Evaluation approaches like nearest-neighbor analysis between synthetic and real datasets were proposed.

\subsubsection{Domain Knowledge Integration}
Clinical priors and structured representations (e.g., anatomical knowledge, temporal models, graph-based ontologies) were cited as useful for guiding models in data-limited scenarios. However, participants cautioned that rigid incorporation of such priors could constrain generalizability. The balance between inductive bias and model flexibility was a recurring theme, especially in highly heterogeneous clinical environments.

\subsubsection{Federated and Collaborative Learning}
Federated Learning (FL) was extensively discussed as a decentralized modeling approach that avoids raw data transfer and supports collaboration across institutions. Participants emphasized its potential but also highlighted several challenges. These included inconsistent data formats, institutional privacy constraints, divergent ontologies, and technical fragility. In some cases, FL resulted in degraded model performance due to mismatches in local data collection protocols and variability in feature and label definitions. Such inconsistencies can reduce the incentive for institutions to participate.
A significant issue raised was the lack of standardization in data de-identification pipelines across health systems. Even leading institutions such as Stanford and UCSF apply different deidentification algorithms, which leads to inconsistent artifacts, breaks in co-reference resolution, and varied replacements for protected health information. These differences hinder cross-site fine-tuning and validation, and can increase the risk of re-identification, especially in rare clinical scenarios.

Despite these limitations, FL was seen as a promising approach if supported by robust schema reconciliation, shared validation strategies, and improved deidentification consistency. Participants also raised an open question: should the research community focus on enhancing federated methods, or prioritize centralizing data using improved deidentification tools and supportive policy measures?

\subsubsection{Evaluation and Benchmarking}
 Establishing reproducibility in small-data environments is difficult without public test sets. The group emphasized the importance of synthetic benchmarking environments, “red team” examples for robustness testing, and probabilistic reporting to quantify uncertainty. They agreed that rigorous evaluation frameworks are needed before small data models can be trusted in critical healthcare workflows.

\subsubsection{Institutional Barriers and Policy}
In addition to technical hurdles, participants identified policy constraints and institutional inertia as significant obstacles to collaborative progress. Trust remains limited across institutions, and ambiguous data stewardship policies often lead to risk-averse behavior. Competitive disincentives also reduce the willingness to share information, including even basic metadata, as many institutions view data as a strategic asset.

The group discussed potential strategies to overcome these barriers. These included organizing hackathons and community challenges, and investing in reproducible synthetic datasets that reflect the structure and statistical properties of real-world data. One specific suggestion was for data holders to release synthetic data in a format that mimics the structure of actual electronic health records. This would allow external teams to prototype models, with final evaluations performed internally on real data by the data custodians.
Participants also noted the potential of platforms like CareEverywhere to support cross-institutional data access. While CareEverywhere enables clinicians to retrieve records from external sites, research reuse remains limited unless the data are explicitly copied into the local chart. This constraint restricts its utility for large-scale research. It highlights the need for clearer pathways that allow for secondary use of such exchanged records in a secure and compliant manner.

\subsection{Conclusion}
The path to scalable AI in healthcare will not be paved by ever-larger datasets alone. Instead, it will require rethinking how we generate, share, and evaluate data and models in low-resource contexts. Our roundtable highlighted that while solutions like synthetic data and federated learning hold great promise, they are not standalone fixes. Their success depends not only on technical rigor but also on policy reform, institutional trust, and a sustained focus on equity and generalizability.

Crucially, participants emphasized that learning from small data is not a temporary workaround—it is a long-term frontier in clinical machine learning. By embedding domain knowledge, quantifying uncertainty, and building collaborative frameworks that respect institutional and regulatory boundaries, we can move toward models that are both practical and principled. The goal is not just to scale AI, but to scale trustworthy AI, especially for the populations and conditions that large datasets often overlook.

\section{Multimodal methods}
\label{sec:table8}

\paragraph{Subtopic:} How should researchers approach multimodal methods in healthcare AI? What are the conceptual and technical challenges in integrating diverse data modalities? How do clinical and computational definitions of “multimodal” diverge, and how does that affect research? 

\paragraph{Chairs:} Jason Fries and Bill Chen 

\paragraph{Participants:} Suhana Bedi, Vasiliki Bikia, Shaza Elsharief, Judy Gichoya, Leeor Hershkovich, Bhavya Jain, Yuxuan Jin, Behdokht Kiafar, Annika Kumar, Daeun Kyung, Simon Lee, Zhongyuan Liang, Chris Lunt, Emma Rocheteau, Kamile Stankeviciute, and Jeremy Weiss

\subsection{Background} In the evolving landscape of smart healthcare, multimodal methods have emerged as a transformative approach, enabling a more comprehensive understanding of patient health. This involves integrating diverse data modalities such as Electronic Health Records (EHRs), medical imaging (e.g., X-rays, CT, MRI), wearable device data, genomic information, environmental factors, and behavioral insights \citep{shaik2024}. Each modality contributes unique insights, and their fusion significantly enhances analysis accuracy and completeness by revealing hidden patterns, correlations, and relationships. This holistic view can be beneficial for optimizing treatment strategies, predicting disease progression, identifying risk factors, and implementing preventive measures, mirroring the complex decision-making processes of clinicians \citep{krones2025}. Studies consistently show that multimodal models outperform single-modality approaches across various performance metrics \citep{teoh2024}. Additionally, recent advancements in foundation models—large-scale, pre-trained AI architectures capable of generalizing across tasks—further propel this field by offering adaptable, multi-purpose solutions capable of handling a broad spectrum of medical tasks and data types, including text, images, and structured information \citep{krones2025, teoh2024}. 

Despite their significant promise, multimodal methods in healthcare AI encounter several challenges. These include ensuring data quality and interoperability across disparate systems and modalities, which is complex and time-consuming due to heterogeneity across populations, regions, and medical centers, and are prone to shifts like concept drift \citep{shaik2024}. Safeguarding patient privacy is critical, especially given the risk of accidental exposure with large foundation models during training \citep{he2025}. Effective data processing and analysis are challenged by the sheer volume and structural differences between modalities, making universal integration and handling missing modalities particularly complex \citep{farhadizadeh2025}. Clinical integration and adoption are hampered by the need for seamless workflow alignment, user-friendly interfaces, adequate training, and amplified regulatory hurdles \citep{soenksen2022}.

\subsection{Discussion} This roundtable brought together an interdisciplinary group of participants, including computer scientists, AI practitioners, and clinicians, to explore current challenges and opportunities in the development and deployment of multimodal methods in healthcare. Given the relevance of large-scale pretrained models to this topic, the conversation naturally overlapped in places with ongoing discussions around foundation models, though the focus remained on the conceptual and practical landscape of multimodal learning. 

Several key themes emerged from the discussion:

\subsubsection{Diverging Definitions and the Importance of Shared Language}
One of the foundational issues raised was the lack of consensus on the definition of “multimodal”. Clinicians often consider imaging and its associated report, such as a radiology scan and its narrative interpretation, as part of a single diagnostic process. In contrast, computer scientists tend to treat these as distinct modalities due to the fundamentally different data types involved. This disciplinary disconnect in terminology and conceptual framing can hinder collaboration and lead to misalignment in system design. Participants emphasized the need for clearer, shared language, especially in interdisciplinary settings where both clinical and computational expertise must align.

Some AI practitioners suggested that the process of tokenization may serve as a practical way to define modalities from a computational standpoint. For example, because images, clinical notes, and time-series signals each require distinct tokenization strategies, they may be considered separate modalities even when clinically intertwined. While this perspective may not be intuitive to clinicians, it reflects an underlying assumption in model design and training pipelines. Acknowledging and bridging this gap will be critical to developing multimodal systems that are both technically robust and clinically meaningful.

\subsubsection{Unified Models vs. Specialized Pipelines}
The group debated whether the future of healthcare AI lies in building a single, flexible “mega-model” capable of reasoning across modalities or in developing more specialized, task-specific models. While the ambition of a unified system is attractive, many noted that current limitations in compute resources, data standardization, and trustworthiness argue for a more incremental approach. Specialized models that work well within defined clinical contexts may be more feasible and immediately impactful. That said, participants also pointed to the potential benefits of richer context integration—such as capturing the interplay between a patient’s lab values, time-series vitals, and narrative notes—which could be lost in siloed models. There was a strong interest in future research that explores ways to link specialized models through a shared representation space or flexible interface.

\subsubsection{Deployment Barriers and Workflow Mismatch}
Participants noted that many multimodal models are developed in isolation from the clinical environments in which they are ultimately intended to be used. A key concern was the misalignment between how and when multimodal data is captured and when clinical decisions are actually made. For example, models trained on fully documented patient timelines often assume access to complete information, whereas real-time clinical settings typically involve partial, evolving, and asynchronous data streams. Clinicians also expressed skepticism about the practical deployment of such systems, raising questions about where the models would run, how their outputs would be delivered during high-stakes interactions like surgeries or primary care visits, and whether the additional cognitive or technical burden would be manageable. These concerns highlighted the need for multimodal research that explicitly considers the deployment context from the outset, rather than treating it as a secondary implementation detail.

\subsubsection{Legal, Ethical, and Governance Considerations}
As multimodal models increasingly incorporate sensitive data types such as imaging, voice recordings, and longitudinal records, concerns around privacy, data ownership, and liability are becoming more prominent. Clinicians emphasized the difficulty of truly de-identifying certain modalities and underscored the need to engage patients transparently in decisions about data governance. The discussion also raised the issue of returning value to data contributors. If patients are providing their data to support the development of AI tools for public benefit, what do they receive in return? While multimodal models offer the promise of population-level improvements, their current opacity and uneven deployment raise important ethical questions about fairness, reciprocity, and the responsible stewardship of patient data.

\subsubsection{The Data Bottleneck}
While methodological innovation continues rapidly, the conversation returns repeatedly to the limitations imposed by the quality, structure, and availability of data. Participants emphasized that medical data is not only sparse and heterogeneous, but also constantly evolving. For example, the quality of MRI scans from the year 2000 differs substantially from those collected today, raising questions about the long-term viability of foundation or multimodal models trained on static datasets.

Some participants argued that we must rethink the data pipeline itself by moving beyond just collecting large quantities toward designing datasets that are well-structured, clinically meaningful, and interoperable. Others suggested formalizing data-focused roles in research teams, akin to software engineering or data stewardship positions, to ensure that data quality and governance keep pace with algorithmic advances.

\section{Scalable and Translational Healthcare Solutions}
\label{sec:table9}

\paragraph{Subtopic:} What are the primary roadblocks preventing promising AI research from becoming widely adopted, impactful clinical tools? What collaborative frameworks (involving academia, industry, clinicians, policymakers, and patients) are needed to accelerate the journey from AI innovation to scalable, translational impact in healthcare? Beyond initial validation, what types of evidence (e.g., pragmatic trials, real-world effectiveness studies, health economic outcomes) are most crucial for convincing healthcare systems to invest in and scale AI solutions? How can we generate this evidence more efficiently?

\paragraph{Chairs:} Niharika S. DSouza and Yixing Jiang

\paragraph{Participants:} Amelia Archer, Josh Fessel, Scott Fleming, Leeor Hershkovich, Chris Lunt, Connor O'Brien, Avishaan Sethi, Shengpu Tang, Aaron Tierney, and Peter Washington

\subsection{Background} Scaling and translating healthcare solutions from the lab to clinical practice is a complex and multifaceted challenge in healthcare research. Successfully bridging the gap between scientific discovery and practical application requires a careful balance of innovation and user-centered design. Translational efforts aim to convert laboratory findings into real-world treatments through collaboration among researchers, clinicians, and industry partners, while maintaining cost-effectiveness. In the face of global health challenges and the drive to speed up the adoption of new therapies and support personalized, evidence-based care, significant obstacles remain—such as protecting data privacy, navigating regulatory hurdles, integrating systems across institutions, and establishing frameworks for user-focused design.
At the CHIL conference, experts from a wide range of backgrounds and experiences shared their insights, offering nuanced perspectives on many facets of this important topic.

\subsection{Discussion}

\subsubsection{Barriers to Translating AI Research into Clinical Practice} One of the primary roadblocks in this space is due to the current disconnect between what researchers consider an advancement vs what is practical in a high-stakes clinical setting with (typically over-worked) human care-givers, such as in an ICU. In developing healthcare solutions that can be adopted in a clinic, two key conditions must be addressed: first, convincing clinicians (eg. nurses, staff) to change their workflow, even minimally, and second, ensuring seamless integration into existing systems without unnecessarily adding new devices or software platforms. It is crucial to understand that each change, such as a new variable to track, or an additional alert, can be quite disruptive in a high-stress situation.

Beyond technological issues, human barriers, such as the challenge of incorporating predictive data into workflows and infrastructure, hinder adoption more often. Unlike the mindset in (machine learning) research, it is imperative to understand solutions are not just about generating more (accurate) predictions but enabling meaningful actions. For example, can a care plan truly change based on a new score that is being reported? In high-intensity settings like the ICU, real-time data risks contributing to information overload and alarm fatigue. Effective design must begin with end users, focusing on actionable outcomes and working backward to shape the solution. Misalignment often arises when researchers prioritize metrics they find important, rather than addressing the practical needs of clinicians and healthcare systems. 

\subsubsection{Collaborative Frameworks}
Nurses, despite being on the frontline of care, are often excluded from key discussions in conferences and collaborative frameworks. Effective healthcare innovation requires participatory design that includes all stakeholders across the pipeline. Human-Computer Interaction (HCI) principles, which emphasize user-centered design, usability, and contextual awareness, are not yet widely integrated into the field of clinical informatics. This gap often leads to tools and systems that are technically robust but misaligned with the actual workflows, cognitive load, and real-world needs of clinicians and patients. A practical step toward bridging this divide is to actively involve HCI experts in clinical informatics forums and conferences—such as inviting them to CHIL—to foster cross-disciplinary dialogue and collaboration. Additionally, patient voices are frequently absent from these conversations, despite being directly impacted by the systems being developed. Including patient representatives in design and evaluation processes is essential for building tools that support not only clinical effectiveness but also patient-centered care.

There's a shift needed from focusing on benchmarks to real-world, in-situ deployment. Additionally, the dissemination of information must be context-aware, tailored to varying levels of technical and clinical expertise. Organizational misalignment—such as disconnects between IT and data science teams—can severely hinder progress, as seen in cases where it took months just to audit and manage even existing models deployed into a single well-funded healthcare system

\subsubsection{Evidence for Clinical AI Deployment}
Fairness in healthcare AI should be studied within the specific deployment context, rather than aiming for universal applicability. Gathering real-world data in these settings helps address practical trade-offs, such as time and patient recruitment, and informs when evidence is sufficient. A bottom-up approach—starting with questions from clinicians rather than forcing existing AI solutions—tends to yield more relevant and useful outcomes, especially since not all problems require AI. Finally, trust in AI is often influenced by hype and marketing, particularly around large language models, which may not always deliver meaningful utility in clinical practice.

\subsubsection{Economic Incentives for Clinical AI}
The impressive demonstration of AI systems in the Chat-GPT era has renewed interest in systems research within healthcare. This is largely driven by the perception that solutions can move quickly from development to deployment, as seen in other application areas. While this attention and momentum is valuable, aligning incentives across stakeholders remains more challenging than ever. Often, efforts focus on specific groups where economic incentives naturally align, but when they don’t, an empathy-driven approach becomes essential to ensure that care and innovation remain inclusive and patient-centered. This might be largely missing from the current efforts being made to map this landscape.

\subsubsection{Ensuring Safe, Ethical, and Adaptive Use at Scale}
As AI and machine learning models become increasingly deployed in healthcare settings, a critical challenge emerges: what happens after deployment? One major concern is the decommissioning of outdated or underperforming models. Healthcare systems often struggle to stop or retire existing tools, leading to bloated infrastructures that become unsustainable. Continuous layering of new tools without retiring old ones can compromise both efficiency and safety.

To address this, robust surveillance and governance systems are essential. These systems must ensure that changes in clinical protocols or reporting standards are communicated to all relevant stakeholders, including data scientists. Currently, lack of communication leads to wasted effort and downstream issues when tools become misaligned with updated practices. For example, a researcher shared months of delay in their analysis and experiments, as their team was not notified when the protocol of reporting a specific clinical measurement was changed abruptly. Integrating data scientists as active members of clinical teams—rather than external consultants—may help maintain alignment and responsiveness throughout the system's lifecycle.

The complexity of modern healthcare IT ecosystems, where multiple systems interface and share data, further complicates governance. Without centralized tracking and clear deployment oversight, duplication of tools and errors can occur. While enterprise-level security controls are crucial, they can also inadvertently become barriers to collaboration and visibility. User leaderboards or similar feedback mechanisms could help capture common tasks and behaviors, allowing teams to prioritize and tailor tool development accordingly.

Finally, the balance between centralized control and distributed innovation remains a tension. Timely interventions—such as those required in trauma resuscitation or early sepsis detection—highlight the importance of deploying systems that are not only timely but also appropriately scoped. Early warning systems have shown promise, but their real-world utility can be undermined by alert fatigue. Instead of promoting tools as fully automated solutions, there’s a growing need to improve training for clinicians and research staff to understand what is a useful measure to compute and report, how-often should one report it, when and how to intervene based on these tools etc, emphasizing partnership over automation.

\section{Summary and Lessons Learned}
Major themes derived from the \nameref{sec:table1&2} roundtable included the importance of user-centric explainability approaches to allow patients and clinicians to meaningfully engage with such tools; achieving this, however, requires tailored development of systems that impose appropriate cognitive load burdens, allowing them to integrate seamlessly into existing workflows. In addition, important concerns regarding potential limitations (e.g., mis-explanations) must be addressed with more infrastructural and regulatory scrutiny and broader education. Further, time-series and sequential data/models introduce an added layer of complexity (beyond typical SHAP/LIME attribution methods) due to the need to explain temporal phenomena— how patient trajectories progress over time. 

At the \nameref{sec:table4} roundtable, discussion first focused on the novelty of building directed acyclic graphs (DAGs) from a combination of domain expert knowledge as well as large language model (LLM) querying, then transitioned to methods to  assess the safety and effectiveness of causal models, and finally concluded by connecting causal inference with reinforcement learning, addressing the role of randomized clinical trials/observational studies in this area, and finally enhancing the generalizability of such approaches through incorporation of economics/econometrics literature.

The \nameref{sec:table3} roundtable emphasized the need to remain acutely aware of the task/problem being solved as well as the problem setting/physical location in which models are deployed, particularly to ensure the inclusion of resource-diverse datasets and populations. The perspectives of clinicians and patients were viewed as equally essential. Ongoing efforts towards addressing the challenge of bias mitigation/retaining fairness, especially for LLMs, were discussed.  

The \nameref{sec:table5} roundtable highlighted limitations in demographic, geographic, and clinical diversity of datasets as a key challenge inhibiting the development of generalizable models. In response, potential state-of-the-art solutions were proposed, including employing federated learning approaches, establishing secure data-pooling frameworks from multiple institutions, and strategies for synthetic data generation. On the algorithmic side, this table reviewed the importance of leveraging active learning, instruction tuning, and tailoring large language models to express uncertainty in their answers to remain robust to data distribution shift with time.  

On the topic of \nameref{sec:table6}, participants discussed the growing integration of LLMs/foundation models into EHR systems for brainstorming medical solutions, thereby serving as assistants to augment the typical established clinical routines. This trend underscores the need for continuous monitoring of these models to accommodate shifting dataset characteristics. Looking ahead, the group anticipated that multi-modal `model composition agents' may shape the next generation of foundation model research and applications.

The roundtable on \nameref{sec:table7} emphasized that while practicing synthetic data generation, federated learning, and domain knowledge integration are essential for data augmentation, structural changes (i.e., policy and institutional/regulatory frameworks) will be necessary to drive progress toward trustworthy AI in the face of the ever-present challenge of small medical data. 

In the realm of \nameref{sec:table8}, issues raised included the importance of definitions of multmodality, the future of unified vs. specialized models, and the role of legal/governance barriers, as well as small-data bottlenecks, overlapping with themes raised in the \nameref{sec:table7} roundtable. 

Finally, the discussion on \nameref{sec:table9} emphasized the importance of the awareness of clinical workflows into which AI will be integrated to optimize its human-computer interaction readiness. Awareness of real-world clinical workflows can be achieved by involving nurses and clinicians earlier in the translation process (e.g., inviting nurses to CHIL!) and integrating data scientists as active members of clinical teams rather than as external consultants. Economic incentives and widespread outreach on successful clinical AI deployment use-cases will also expedite the translation of trustworthy AI to the clinic.

Overall, our CHIL 2025 roundtables were well-attended and enjoyed by participants, roundtable chairs, and organizing committee members.  They served to facilitate discussions about pertinent topics to our AI-in-healthcare community, foster connections between medical practitioners, scientists, researchers, and policy-makers, and enabled the written compilation of key challenges, advances, and outlooks for the future across the above 8 research subtopics.  While the format, discussion length, and organizational structure (pre- and post-material preparations) were appreciated by our attendees and chairs, one suggestion for the future may be to more-strictly enforce transitions between roundtables after the 30-minute discussion period to enable more exposure to different subtopics among more participants.  This could be carried out by an audible buzzer being played after each discussion time-interval or possibly by shortening even further the discussion time at each roundtable from 30 minutes to 15 minutes.  We look forward to the continuation of Research Roundtables and to the stimulating conversations to-come at CHIL 2026!

\acks{We would like to sincerely thank the senior and junior chairs of each roundtable for their leadership in facilitating thoughtful, inclusive discussions and for compiling the insightful summaries featured in this document. We are also grateful to the CHIL 2025 General Chairs, Matthew McDermott and Irene Chen, for their guidance and for reviewing this compilation.  We also would like to thank CHIL 2025 Program Chairs Jessilyn Dunn and Roxana Daneshjou for their guidance and roundtable chair recommendations. Special thanks to Xuhai `Orson' Xu for help in determining authors of submitted/accepted CHIL papers and to the entire CHIL 2025 Organizing Committee.  Finally, our gratitude goes to Tasmie Sarker and Daseyra Fontalvo for all of their organizational and logistical support which enabled the success our CHIL 2025 roundtables. Finally, our thanks go to the board members of the Association for Health, Inference, and Learning.}

\bibliography{chil25_roundtables}

\begin{thebibliography}{41}
\providecommand{\natexlab}[1]{#1}
\providecommand{\url}[1]{\texttt{#1}}
\expandafter\ifx\csname urlstyle\endcsname\relax
  \providecommand{\doi}[1]{doi: #1}\else
  \providecommand{\doi}{doi: \begingroup \urlstyle{rm}\Url}\fi

\bibitem[Ao et~al.(2025)Ao, Chen, Li, Nie, Zhang, and Chen]{ao2025}
Guangyu Ao, Min Chen, Jing Li, Huibing Nie, Lei Zhang, and Zejun Chen.
\newblock Comparative analysis of large language models on rare disease identification.
\newblock \emph{Orphanet Journal of Rare Diseases}, 20:\penalty0 150, April 2025.
\newblock ISSN 1750-1172.
\newblock \doi{10.1186/s13023-025-03656-w}.
\newblock URL \url{https://www.ncbi.nlm.nih.gov/pmc/articles/PMC11959745/}.

\bibitem[Arora et~al.(2023)Arora, Alderman, Palmer, Ganapathi, Laws, McCradden, Oakden-Rayner, Pfohl, Ghassemi, McKay, Treanor, Rostamzadeh, Mateen, Gath, Adebajo, Kuku, Matin, Heller, Sapey, Sebire, Cole-Lewis, Calvert, Denniston, and Liu]{arora2023}
Anmol Arora, Joseph~E. Alderman, Joanne Palmer, Shaswath Ganapathi, Elinor Laws, Melissa~D. McCradden, Lauren Oakden-Rayner, Stephen~R. Pfohl, Marzyeh Ghassemi, Francis McKay, Darren Treanor, Negar Rostamzadeh, Bilal Mateen, Jacqui Gath, Adewole~O. Adebajo, Stephanie Kuku, Rubeta Matin, Katherine Heller, Elizabeth Sapey, Neil~J. Sebire, Heather Cole-Lewis, Melanie Calvert, Alastair Denniston, and Xiaoxuan Liu.
\newblock The value of standards for health datasets in artificial intelligence-based applications.
\newblock \emph{Nature Medicine}, 29\penalty0 (11):\penalty0 2929--2938, 2023.
\newblock ISSN 1078-8956.
\newblock \doi{10.1038/s41591-023-02608-w}.
\newblock URL \url{https://www.ncbi.nlm.nih.gov/pmc/articles/PMC10667100/}.

\bibitem[Bedi et~al.(2025)Bedi, Liu, Orr-Ewing, Dash, Koyejo, Callahan, Fries, Wornow, Swaminathan, Lehmann, Hong, Kashyap, Chaurasia, Shah, Singh, Tazbaz, Milstein, Pfeffer, and Shah]{bedi2025}
Suhana Bedi, Yutong Liu, Lucy Orr-Ewing, Dev Dash, Sanmi Koyejo, Alison Callahan, Jason~A. Fries, Michael Wornow, Akshay Swaminathan, Lisa~Soleymani Lehmann, Hyo~Jung Hong, Mehr Kashyap, Akash~R. Chaurasia, Nirav~R. Shah, Karandeep Singh, Troy Tazbaz, Arnold Milstein, Michael~A. Pfeffer, and Nigam~H. Shah.
\newblock Testing and {Evaluation} of {Health} {Care} {Applications} of {Large} {Language} {Models}.
\newblock \emph{JAMA}, 333\penalty0 (4):\penalty0 319--328, January 2025.
\newblock ISSN 0098-7484.
\newblock \doi{10.1001/jama.2024.21700}.
\newblock URL \url{https://www.ncbi.nlm.nih.gov/pmc/articles/PMC11480901/}.

\bibitem[Bewersdorff et~al.(2025)Bewersdorff, Hornberger, Nerdel, and Schiff]{bewersdorff2025}
Arne Bewersdorff, Marie Hornberger, Claudia Nerdel, and Daniel~S. Schiff.
\newblock {AI} advocates and cautious critics: {How} {AI} attitudes, {AI} interest, use of {AI}, and {AI} literacy build university students' {AI} self-efficacy.
\newblock \emph{Computers and Education: Artificial Intelligence}, 8:\penalty0 100340, June 2025.
\newblock ISSN 2666-920X.
\newblock \doi{10.1016/j.caeai.2024.100340}.
\newblock URL \url{https://www.sciencedirect.com/science/article/pii/S2666920X24001437}.

\bibitem[Boughdiri et~al.(2025)Boughdiri, Berenfeld, Josse, and Scornet]{boughdiri2025}
Ahmed Boughdiri, Clément Berenfeld, Julie Josse, and Erwan Scornet.
\newblock A {Unified} {Framework} for the {Transportability} of {Population}-{Level} {Causal} {Measures}, May 2025.
\newblock URL \url{http://arxiv.org/abs/2505.13104}.
\newblock arXiv:2505.13104 [stat].

\bibitem[Chen et~al.(2021)Chen, Pierson, Rose, Joshi, Ferryman, and Ghassemi]{chen2021}
Irene~Y. Chen, Emma Pierson, Sherri Rose, Shalmali Joshi, Kadija Ferryman, and Marzyeh Ghassemi.
\newblock Ethical {Machine} {Learning} in {Healthcare}.
\newblock \emph{Annual review of biomedical data science}, 4:\penalty0 123--144, July 2021.
\newblock ISSN 2574-3414.
\newblock \doi{10.1146/annurev-biodatasci-092820-114757}.
\newblock URL \url{https://www.ncbi.nlm.nih.gov/pmc/articles/PMC8362902/}.

\bibitem[Chiang et~al.(2025)Chiang, Shanmugam, Beecy, Sayer, Estrin, Garg, and Pierson]{chiang2025}
Erica Chiang, Divya Shanmugam, Ashley~N. Beecy, Gabriel Sayer, Deborah Estrin, Nikhil Garg, and Emma Pierson.
\newblock Learning {Disease} {Progression} {Models} {That} {Capture} {Health} {Disparities}, April 2025.
\newblock URL \url{http://arxiv.org/abs/2412.16406}.
\newblock arXiv:2412.16406 [cs].

\bibitem[Chouldechova(2016)]{chouldechova2016}
Alexandra Chouldechova.
\newblock Fair prediction with disparate impact: {A} study of bias in recidivism prediction instruments, October 2016.
\newblock URL \url{http://arxiv.org/abs/1610.07524}.
\newblock arXiv:1610.07524 [stat].

\bibitem[Colnet et~al.(2023)Colnet, Mayer, Chen, Dieng, Li, Varoquaux, Vert, Josse, and Yang]{colnet2023}
Bénédicte Colnet, Imke Mayer, Guanhua Chen, Awa Dieng, Ruohong Li, Gaël Varoquaux, Jean-Philippe Vert, Julie Josse, and Shu Yang.
\newblock Causal inference methods for combining randomized trials and observational studies: a review, January 2023.
\newblock URL \url{http://arxiv.org/abs/2011.08047}.
\newblock arXiv:2011.08047 [stat].

\bibitem[Colnet et~al.(2024)Colnet, Josse, Varoquaux, and Scornet]{colnet2024}
Bénédicte Colnet, Julie Josse, Gaël Varoquaux, and Erwan Scornet.
\newblock Risk ratio, odds ratio, risk difference... {Which} causal measure is easier to generalize?, March 2024.
\newblock URL \url{http://arxiv.org/abs/2303.16008}.
\newblock arXiv:2303.16008 [stat].

\bibitem[Dahabreh et~al.(2024)Dahabreh, Matthews, Steingrimsson, Scharfstein, and Stuart]{dahabreh2024}
Issa~J Dahabreh, Anthony Matthews, Jon~A Steingrimsson, Daniel~O Scharfstein, and Elizabeth~A Stuart.
\newblock Using {Trial} and {Observational} {Data} to {Assess} {Effectiveness}: {Trial} {Emulation}, {Transportability}, {Benchmarking}, and {Joint} {Analysis}.
\newblock \emph{Epidemiologic Reviews}, 46\penalty0 (1):\penalty0 1--16, September 2024.
\newblock ISSN 1478-6729.
\newblock \doi{10.1093/epirev/mxac011}.
\newblock URL \url{https://doi.org/10.1093/epirev/mxac011}.

\bibitem[Degtiar and Rose(2023)]{degtiar2023}
Irina Degtiar and Sherri Rose.
\newblock A {Review} of {Generalizability} and {Transportability}.
\newblock \emph{Annual Review of Statistics and Its Application}, 10\penalty0 (1):\penalty0 501--524, March 2023.
\newblock ISSN 2326-8298, 2326-831X.
\newblock \doi{10.1146/annurev-statistics-042522-103837}.
\newblock URL \url{https://www.annualreviews.org/doi/10.1146/annurev-statistics-042522-103837}.
\newblock Publisher: Annual Reviews.

\bibitem[Even and Josse(2025)]{even2025}
Mathieu Even and Julie Josse.
\newblock Rethinking the {Win} {Ratio}: {A} {Causal} {Framework} for {Hierarchical} {Outcome} {Analysis}, April 2025.
\newblock URL \url{http://arxiv.org/abs/2501.16933}.
\newblock arXiv:2501.16933 [stat].

\bibitem[Farhadizadeh et~al.(2025)Farhadizadeh, Weymann, Blaß, Kraus, Gundler, Walter, Hempen, Binder, and Binder]{farhadizadeh2025}
Maryam Farhadizadeh, Maria Weymann, Michael Blaß, Johann Kraus, Christopher Gundler, Sebastian Walter, Noah Hempen, Harald Binder, and Nadine Binder.
\newblock A systematic review of challenges and proposed solutions in modeling multimodal data, May 2025.
\newblock URL \url{http://arxiv.org/abs/2505.06945}.
\newblock arXiv:2505.06945 [cs].

\bibitem[Grant and Wood(2022)]{grant2022}
Torie~L. Grant and Robert~A. Wood.
\newblock The influence of urban exposures and residence on childhood asthma.
\newblock \emph{Pediatric Allergy and Immunology}, 33\penalty0 (5):\penalty0 e13784, May 2022.
\newblock ISSN 0905-6157.
\newblock \doi{10.1111/pai.13784}.
\newblock URL \url{https://www.ncbi.nlm.nih.gov/pmc/articles/PMC9288815/}.

\bibitem[He et~al.(2025)He, Huang, Jiang, Nie, Wang, Wang, and Chen]{he2025}
Yuting He, Fuxiang Huang, Xinrui Jiang, Yuxiang Nie, Minghao Wang, Jiguang Wang, and Hao Chen.
\newblock Foundation {Model} for {Advancing} {Healthcare}: {Challenges}, {Opportunities} and {Future} {Directions}.
\newblock \emph{IEEE Reviews in Biomedical Engineering}, 18:\penalty0 172--191, 2025.
\newblock ISSN 1941-1189.
\newblock \doi{10.1109/RBME.2024.3496744}.
\newblock URL \url{https://ieeexplore.ieee.org/document/10750441}.

\bibitem[Huang and Parikh(2024)]{huang2024a}
Melody Huang and Harsh Parikh.
\newblock Toward {Generalizing} {Inferences} {From} {Trials} to {Target} {Populations}.
\newblock \emph{Harvard Data Science Review}, August 2024.
\newblock \doi{10.1162/99608f92.68c7973b}.
\newblock URL \url{https://hdsr.mitpress.mit.edu/pub/pzkwyen2/release/2}.
\newblock Publisher: The MIT Press.

\bibitem[{IBM Research}(2017)]{ibmresearch2017}
{IBM Research}.
\newblock {IBM} researchers bring {AI} to radiology, June 2017.
\newblock URL \url{https://www.youtube.com/watch?v=XLb0xUe80uo}.

\bibitem[Kleinberg et~al.(2016)Kleinberg, Mullainathan, and Raghavan]{kleinberg2016}
Jon Kleinberg, Sendhil Mullainathan, and Manish Raghavan.
\newblock Inherent {Trade}-{Offs} in the {Fair} {Determination} of {Risk} {Scores}, November 2016.
\newblock URL \url{http://arxiv.org/abs/1609.05807}.
\newblock arXiv:1609.05807 [cs].

\bibitem[Krones et~al.(2025)Krones, Marikkar, Parsons, Szmul, and Mahdi]{krones2025}
Felix Krones, Umar Marikkar, Guy Parsons, Adam Szmul, and Adam Mahdi.
\newblock Review of multimodal machine learning approaches in healthcare.
\newblock \emph{Information Fusion}, 114:\penalty0 102690, February 2025.
\newblock ISSN 1566-2535.
\newblock \doi{10.1016/j.inffus.2024.102690}.
\newblock URL \url{https://www.sciencedirect.com/science/article/pii/S1566253524004688}.

\bibitem[Lundberg and Lee(2017)]{lundberg2017}
Scott~M Lundberg and Su-In Lee.
\newblock A {Unified} {Approach} to {Interpreting} {Model} {Predictions}.
\newblock In \emph{Advances in {Neural} {Information} {Processing} {Systems}}, volume~30. Curran Associates, Inc., 2017.
\newblock URL \url{https://proceedings.neurips.cc/paper/2017/hash/8a20a8621978632d76c43dfd28b67767-Abstract.html}.

\bibitem[López et~al.(2025)López, Elsharief, Jorf, Darwish, Ma, and Shamout]{lopez2025}
L.~Julián~Lechuga López, Shaza Elsharief, Dhiyaa~Al Jorf, Firas Darwish, Congbo Ma, and Farah~E. Shamout.
\newblock Uncertainty {Quantification} for {Machine} {Learning} in {Healthcare}: {A} {Survey}, May 2025.
\newblock URL \url{http://arxiv.org/abs/2505.02874}.
\newblock arXiv:2505.02874 [cs].

\bibitem[Manke-Reimers et~al.(2025)Manke-Reimers, Brugger, Bärnighausen, and Kohler]{manke-reimers2025}
Fabian Manke-Reimers, Vincent Brugger, Till Bärnighausen, and Stefan Kohler.
\newblock When, why and how are estimated effects transported between populations? {A} scoping review of studies applying transportability methods.
\newblock \emph{European Journal of Epidemiology}, 40\penalty0 (3):\penalty0 255--273, April 2025.
\newblock ISSN 1573-7284.
\newblock \doi{10.1007/s10654-025-01217-w}.
\newblock URL \url{https://link.springer.com/article/10.1007/s10654-025-01217-w}.
\newblock Number: 3 Publisher: Springer.

\bibitem[Mehrabi et~al.(2021)Mehrabi, Morstatter, Saxena, Lerman, and Galstyan]{mehrabi2021}
Ninareh Mehrabi, Fred Morstatter, Nripsuta Saxena, Kristina Lerman, and Aram Galstyan.
\newblock A {Survey} on {Bias} and {Fairness} in {Machine} {Learning}.
\newblock \emph{ACM Comput. Surv.}, 54\penalty0 (6):\penalty0 115:1--115:35, July 2021.
\newblock ISSN 0360-0300.
\newblock \doi{10.1145/3457607}.
\newblock URL \url{https://doi.org/10.1145/3457607}.

\bibitem[Miao et~al.(2025)Miao, Williams, Chinedu-Eneh, Zack, Alsentzer, Butte, and Chen]{miao2025a}
Brenda~Y. Miao, Christopher Y.~K. Williams, Ebenezer Chinedu-Eneh, Travis Zack, Emily Alsentzer, Atul~J. Butte, and Irene~Y. Chen.
\newblock Understanding contraceptive switching rationales from real world clinical notes using large language models.
\newblock \emph{npj Digital Medicine}, 8\penalty0 (1):\penalty0 221, April 2025.
\newblock ISSN 2398-6352.
\newblock \doi{10.1038/s41746-025-01615-0}.
\newblock URL \url{https://www.nature.com/articles/s41746-025-01615-0}.
\newblock Publisher: Nature Publishing Group.

\bibitem[Moy et~al.(2024)Moy, Irannejad, Manning, Farahani, Ahmed, Gao, Prabhune, Lorenz, Mirza, and Klinger]{moy2024}
Sally Moy, Mona Irannejad, Stephanie~Jeanneret Manning, Mehrdad Farahani, Yomna Ahmed, Ellis Gao, Radhika Prabhune, Suzan Lorenz, Raza Mirza, and Christopher Klinger.
\newblock Patient {Perspectives} on the {Use} of {Artificial} {Intelligence} in {Health} {Care}: {A} {Scoping} {Review}.
\newblock \emph{Journal of Patient-Centered Research and Reviews}, 11\penalty0 (1):\penalty0 51--62, April 2024.
\newblock ISSN 2330-068X.
\newblock \doi{10.17294/2330-0698.2029}.
\newblock URL \url{https://www.ncbi.nlm.nih.gov/pmc/articles/PMC11000703/}.

\bibitem[Pfohl et~al.(2021)Pfohl, Foryciarz, and Shah]{pfohl2021}
Stephen~R. Pfohl, Agata Foryciarz, and Nigam~H. Shah.
\newblock An empirical characterization of fair machine learning for clinical risk prediction.
\newblock \emph{Journal of Biomedical Informatics}, 113:\penalty0 103621, January 2021.
\newblock ISSN 1532-0464.
\newblock \doi{10.1016/j.jbi.2020.103621}.
\newblock URL \url{https://www.sciencedirect.com/science/article/pii/S1532046420302495}.

\bibitem[Pfohl et~al.(2024)Pfohl, Cole-Lewis, Sayres, Neal, Asiedu, Dieng, Tomasev, Rashid, Azizi, Rostamzadeh, McCoy, Celi, Liu, Schaekermann, Walton, Parrish, Nagpal, Singh, Dewitt, Mansfield, Prakash, Heller, Karthikesalingam, Semturs, Barral, Corrado, Matias, Smith-Loud, Horn, and Singhal]{pfohl2024}
Stephen~R. Pfohl, Heather Cole-Lewis, Rory Sayres, Darlene Neal, Mercy Asiedu, Awa Dieng, Nenad Tomasev, Qazi~Mamunur Rashid, Shekoofeh Azizi, Negar Rostamzadeh, Liam~G. McCoy, Leo~Anthony Celi, Yun Liu, Mike Schaekermann, Alanna Walton, Alicia Parrish, Chirag Nagpal, Preeti Singh, Akeiylah Dewitt, Philip Mansfield, Sushant Prakash, Katherine Heller, Alan Karthikesalingam, Christopher Semturs, Joelle Barral, Greg Corrado, Yossi Matias, Jamila Smith-Loud, Ivor Horn, and Karan Singhal.
\newblock A {Toolbox} for {Surfacing} {Health} {Equity} {Harms} and {Biases} in {Large} {Language} {Models}.
\newblock \emph{Nature Medicine}, 30\penalty0 (12):\penalty0 3590--3600, December 2024.
\newblock ISSN 1078-8956, 1546-170X.
\newblock \doi{10.1038/s41591-024-03258-2}.
\newblock URL \url{http://arxiv.org/abs/2403.12025}.
\newblock arXiv:2403.12025 [cs].

\bibitem[Pfohl et~al.(2025)Pfohl, Harris, Nagpal, Madras, Mhasawade, Salaudeen, Dieng, Sequeira, Arciniegas, Sung, Ezeanochie, Cole-Lewis, Heller, Koyejo, and D'Amour]{pfohl2025}
Stephen~R. Pfohl, Natalie Harris, Chirag Nagpal, David Madras, Vishwali Mhasawade, Olawale Salaudeen, Awa Dieng, Shannon Sequeira, Santiago Arciniegas, Lillian Sung, Nnamdi Ezeanochie, Heather Cole-Lewis, Katherine Heller, Sanmi Koyejo, and Alexander D'Amour.
\newblock Understanding challenges to the interpretation of disaggregated evaluations of algorithmic fairness, June 2025.
\newblock URL \url{http://arxiv.org/abs/2506.04193}.
\newblock arXiv:2506.04193 [stat].

\bibitem[Pierson et~al.(2025)Pierson, Shanmugam, Movva, Kleinberg, Agrawal, Dredze, Ferryman, Gichoya, Jurafsky, Koh, Levy, Mullainathan, Obermeyer, Suresh, and Vafa]{pierson2025}
Emma Pierson, Divya Shanmugam, Rajiv Movva, Jon Kleinberg, Monica Agrawal, Mark Dredze, Kadija Ferryman, Judy~Wawira Gichoya, Dan Jurafsky, Pang~Wei Koh, Karen Levy, Sendhil Mullainathan, Ziad Obermeyer, Harini Suresh, and Keyon Vafa.
\newblock Using {Large} {Language} {Models} to {Promote} {Health} {Equity}.
\newblock \emph{NEJM AI}, 2\penalty0 (2):\penalty0 AIp2400889, January 2025.
\newblock \doi{10.1056/AIp2400889}.
\newblock URL \url{https://ai.nejm.org/doi/full/10.1056/AIp2400889}.
\newblock Publisher: Massachusetts Medical Society.

\bibitem[Ribeiro et~al.(2016)Ribeiro, Singh, and Guestrin]{ribeiro2016}
Marco~Tulio Ribeiro, Sameer Singh, and Carlos Guestrin.
\newblock "{Why} {Should} {I} {Trust} {You}?": {Explaining} the {Predictions} of {Any} {Classifier}.
\newblock In \emph{Proceedings of the 22nd {ACM} {SIGKDD} {International} {Conference} on {Knowledge} {Discovery} and {Data} {Mining}}, {KDD} '16, pages 1135--1144, New York, NY, USA, August 2016. Association for Computing Machinery.
\newblock ISBN 978-1-4503-4232-2.
\newblock \doi{10.1145/2939672.2939778}.
\newblock URL \url{https://dl.acm.org/doi/10.1145/2939672.2939778}.

\bibitem[Rosenman(2025)]{rosenman2025}
Evan T.~R. Rosenman.
\newblock Methods for {Combining} {Observational} and {Experimental} {Causal} {Estimates}: {A} {Review}.
\newblock \emph{WIREs Computational Statistics}, 17\penalty0 (2):\penalty0 e70027, 2025.
\newblock ISSN 1939-0068.
\newblock \doi{10.1002/wics.70027}.
\newblock URL \url{https://onlinelibrary.wiley.com/doi/abs/10.1002/wics.70027}.
\newblock \_eprint: https://wires.onlinelibrary.wiley.com/doi/pdf/10.1002/wics.70027.

\bibitem[Scantamburlo et~al.(2025)Scantamburlo, Baumann, and Heitz]{scantamburlo2025}
Teresa Scantamburlo, Joachim Baumann, and Christoph Heitz.
\newblock On prediction-modelers and decision-makers: why fairness requires more than a fair prediction model.
\newblock \emph{Ai \& Society}, 40\penalty0 (2):\penalty0 353--369, 2025.
\newblock ISSN 0951-5666.
\newblock \doi{10.1007/s00146-024-01886-3}.
\newblock URL \url{https://www.ncbi.nlm.nih.gov/pmc/articles/PMC11968481/}.

\bibitem[Shaik et~al.(2024)Shaik, Tao, Li, Xie, and Velásquez]{shaik2024}
Thanveer Shaik, Xiaohui Tao, Lin Li, Haoran Xie, and Juan~D. Velásquez.
\newblock A survey of multimodal information fusion for smart healthcare: {Mapping} the journey from data to wisdom.
\newblock \emph{Information Fusion}, 102:\penalty0 102040, February 2024.
\newblock ISSN 1566-2535.
\newblock \doi{10.1016/j.inffus.2023.102040}.
\newblock URL \url{https://www.sciencedirect.com/science/article/pii/S1566253523003561}.

\bibitem[Shanmugam et~al.(2023)Shanmugam, Hou, and Pierson]{shanmugam2023}
Divya Shanmugam, Kaihua Hou, and Emma Pierson.
\newblock Quantifying disparities in intimate partner violence: a machine learning method to correct for underreporting, December 2023.
\newblock URL \url{http://arxiv.org/abs/2110.04133}.
\newblock arXiv:2110.04133 [cs].

\bibitem[Soenksen et~al.(2022)Soenksen, Ma, Zeng, Boussioux, Villalobos~Carballo, Na, Wiberg, Li, Fuentes, and Bertsimas]{soenksen2022}
Luis~R. Soenksen, Yu~Ma, Cynthia Zeng, Leonard Boussioux, Kimberly Villalobos~Carballo, Liangyuan Na, Holly~M. Wiberg, Michael~L. Li, Ignacio Fuentes, and Dimitris Bertsimas.
\newblock Integrated multimodal artificial intelligence framework for healthcare applications.
\newblock \emph{npj Digital Medicine}, 5\penalty0 (1):\penalty0 149, September 2022.
\newblock ISSN 2398-6352.
\newblock \doi{10.1038/s41746-022-00689-4}.
\newblock URL \url{https://www.nature.com/articles/s41746-022-00689-4}.
\newblock Publisher: Nature Publishing Group.

\bibitem[Sun et~al.(2023)Sun, Bhave, Altosaar, and Elhadad]{sun2023}
Tony~Y. Sun, Shreyas~A. Bhave, Jaan Altosaar, and Noémie Elhadad.
\newblock Assessing {Phenotype} {Definitions} for {Algorithmic} {Fairness}.
\newblock \emph{AMIA Annual Symposium Proceedings}, 2022:\penalty0 1032--1041, April 2023.
\newblock ISSN 1942-597X.
\newblock URL \url{https://www.ncbi.nlm.nih.gov/pmc/articles/PMC10148336/}.

\bibitem[Teoh et~al.(2024)Teoh, Dong, Zuo, Lai, Hasikin, and Wu]{teoh2024}
Jing~Ru Teoh, Jian Dong, Xiaowei Zuo, Khin~Wee Lai, Khairunnisa Hasikin, and Xiang Wu.
\newblock Advancing healthcare through multimodal data fusion: a comprehensive review of techniques and applications.
\newblock \emph{PeerJ Computer Science}, 10:\penalty0 e2298, October 2024.
\newblock ISSN 2376-5992.
\newblock \doi{10.7717/peerj-cs.2298}.
\newblock URL \url{https://www.ncbi.nlm.nih.gov/pmc/articles/PMC11623190/}.

\bibitem[Tulk~Jesso et~al.(2022)Tulk~Jesso, Kelliher, Sanghavi, Martin, and Henrickson~Parker]{tulkjesso2022}
Stephanie Tulk~Jesso, Aisling Kelliher, Harsh Sanghavi, Thomas Martin, and Sarah Henrickson~Parker.
\newblock Inclusion of {Clinicians} in the {Development} and {Evaluation} of {Clinical} {Artificial} {Intelligence} {Tools}: {A} {Systematic} {Literature} {Review}.
\newblock \emph{Frontiers in Psychology}, 13:\penalty0 830345, April 2022.
\newblock ISSN 1664-1078.
\newblock \doi{10.3389/fpsyg.2022.830345}.
\newblock URL \url{https://www.ncbi.nlm.nih.gov/pmc/articles/PMC9022040/}.

\bibitem[Walsh et~al.(2017)Walsh, Sharman, and Hripcsak]{walsh2017}
Colin~G. Walsh, Kavya Sharman, and George Hripcsak.
\newblock Beyond discrimination: {A} comparison of calibration methods and clinical usefulness of predictive models of readmission risk.
\newblock \emph{Journal of Biomedical Informatics}, 76:\penalty0 9--18, December 2017.
\newblock ISSN 1532-0464.
\newblock \doi{10.1016/j.jbi.2017.10.008}.
\newblock URL \url{https://www.sciencedirect.com/science/article/pii/S1532046417302277}.

\bibitem[Ye et~al.(2024)Ye, Wang, Huang, Chen, Zhang, Moniz, Gao, Geyer, Huang, Chen, Chawla, and Zhang]{ye2025}
Jiayi Ye, Yanbo Wang, Yue Huang, Dongping Chen, Qihui Zhang, Nuno Moniz, Tian Gao, Werner Geyer, Chao Huang, Pin-Yu Chen, Nitesh~V Chawla, and Xiangliang Zhang.
\newblock Justice or {Prejudice}? {Qunatifying} {Biases} in {LLM}-as-a-{Judge}.
\newblock 2024.
\newblock URL \url{https://arxiv.org/abs/2410.02736}.

\end{thebibliography}

\end{document}